\documentclass[10pt, conference, compsocconf]{IEEEtran}

\usepackage{epsfig}
\usepackage{CJK}
\usepackage{algorithm}
\usepackage{algorithmicx}
\usepackage{algpseudocode}
\usepackage{multirow}
\usepackage{setspace}
\usepackage{bibspacing}
\usepackage{amsmath}
\usepackage[marginal]{footmisc}  

\begin{document}
%
\title{Curved Text Detection in Natural Scene Images \\with Semi- and Weakly-Supervised Learning}

\author{
\IEEEauthorblockN{Xugong Qin, Yu Zhou\IEEEauthorrefmark{1}, Dongbao Yang, Weiping Wang\\
Institute of Information Engineering, Chinese Academy of Sciences, Beijing, China\\
School of Cyber Security, University of Chinese Academy of Sciences, Beijing, China\\
Email: \{qinxugong, zhouyu, yangdongbao, wangweiping\}@iie.ac.cn}}

\maketitle

\begin{abstract}
Detecting curved text in the wild is very challenging. Recently, most state-of-the-art methods are segmentation based and require pixel-level annotations.
We propose a novel scheme to train an accurate text detector using only a small amount of pixel-level annotated data and
a large amount of data annotated with rectangles or even unlabeled data. A baseline model is first obtained by training
with the pixel-level annotated data and then used to annotate
unlabeled or weakly labeled data. A novel strategy which utilizes ground-truth bounding boxes to generate pseudo mask annotations is proposed in weakly-supervised learning.
Experimental results on CTW1500 and Total-Text demonstrate that our method can substantially reduce the requirement of pixel-level annotated data.
Our method can also generalize well across two datasets.
The performance of the proposed method is comparable with the state-of-the-art methods with only 10\% pixel-level annotated data and 90\% rectangle-level weakly annotated data.


\end{abstract}


\IEEEpeerreviewmaketitle

\footnote{* The corresponding author}
\section{Introduction}
\label{sec1}


Scene text is one of the most common objects in images, which usually appears on license
plates, product packages, billboards, etc, and carries rich semantic information.
Reading text in natural scene images is a basic task for various tasks like vehicle
auto-navigation and product retrieval, and attracts more and more attentions
from both academic and industrial community. Compared to common objects, scene text{}
is born with multiple orientations, large aspect ratios, arbitrary shapes or layouts
and complex backgrounds, which brings difficulties for detection and recognition\cite{ye2015text}\cite{zhu2016scene}\cite{long2018scene}.

Deep learning methods has dominated the scene text detection areas in recent years.
Accompanied with the development of algorithms, lots of datasets appears including ICDAR2013, ICDAR2015, Total-Text, CTW1500, etc,
and the annotations evolve from coarse to fine.
In ICDAR2013 \cite{karatzas2013icdar}, scene texts are almost horizontal and annotated with horizontal rectangles.
To annotate multi-oriented scene texts more compactly, quadrilaterals are adopted in ICDAR2015 \cite{karatzas2015icdar}.
To spur an interest in the community to address curved text, Total-Text \cite{ch2017total} is proposed.
In this dataset, the authors go a step further and annotate words with polygons of different numbers.
In CTW1500 \cite{yuliang2017detecting}, a much finer strategy is adopted to annotate text lines with 14 points polygons.

Recently, various methods \cite{long2018textsnake}\cite{dai2018fused}\cite{ch2017total}\cite{lyu2018mask} based on semantic
segmentation \cite{long2015fully} and instance segmentation \cite{li2017fully} are
proposed to detect text of arbitrary shape.
They demonstrate superior performance and generalization on various datasets.
However, to achieve such performance, pixel-level annotations are required, which are more laborious and tedious compared with
rectangle, quadrilateral and polygon based annotations. An alternative is to use pixel-level annotations converted from polygons.
However, polygon-based annotations are still more expensive than rectangle based annotations. Therefore, using less amount of pixel-level
annotated data together with large amount of rectangle based annotated data to train an accurate segmentation based detector
is very challenging and of great importance. In this work, we propose a semi- and weakly-supervised framework to solve this problem.


In this framework, a text detector and three strategies are involved to effectively utilize unannotated and
weakly annotated data to boost performance.
We build a network that performs bounding box localization and text/non-text classification and simultaneously
identifies the instance level text masks within the bounding boxes. The text masks are taken as the detection results.
At first, a baseline model is obtained with a small amount of pixel-level annotated data.
Then the model is used to exploit more pseudo annotations from unannotated data or weakly annotated data with horizontal rectangles
by three different strategies.


For semi-supervised learning, a naive strategy identifies box and mask outputs as pseudo annotations through a thresholding process.
For weakly-supervised learning, a filter strategy selects box and mask outputs as pseudo annotations after a filtering process
inspired by WeText \cite{tian2017wetext}.
The other weakly-supervised learning strategy named local, takes boxes as proposals, directly producing pseudo mask annotations.
The pseudo annotated data together with the original annotated data are used to retrain a superior model.
Recursive training in which the predicted outputs are taken as supervision in next training round is adopted in our experiment.
We observe the improvement of performance within several training rounds.

\begin{figure*}[t]
	\centering
	\includegraphics[width=0.9\linewidth]{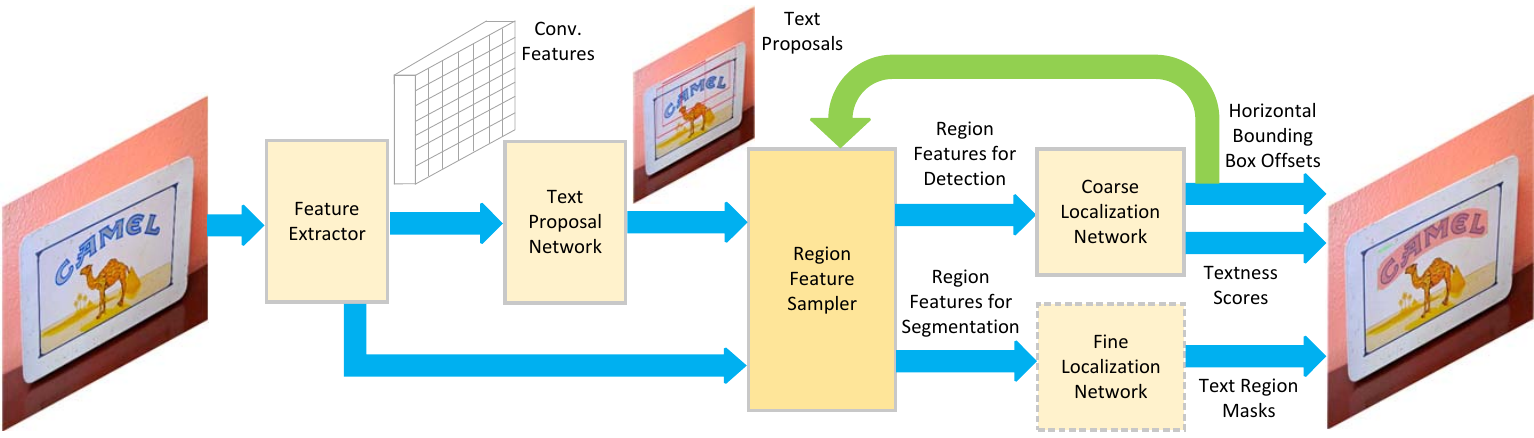}
	\caption{Illustration of the structure of the curved text detector\cite{li2017towards}.}
	\label{struc}
\end{figure*}

The contributions of this work are summarized as below.
\begin{itemize}
\item First, we propose a semi- and weakly-supervised curved text detection framework in which an accurate scene text detector is trained
by utilizing a small amount data with strong polygon/pixel-level annotations and a large amount of unannotated or weakly annotated data.
\item Second, a novel precise pseudo mask annotation generation strategy which utilizes ground-truth bounding boxes as proposals is proposed
in weakly-supervised learning.
\item Third, the experimental results on CTW1500 and Total-Text demonstrate that our method can substantially reduce the requirement of
pixel-level annotations,
and results across the two datasets show good generalization of the proposed method.
The performance of our method is comparable to the state of the art methods
with only 10\% pixel-level annotated training data and 90\% weakly rectangle based annotated training data.
\end{itemize}

\section{Related work}
\label{sec2}
In this section, we review the most relevant works on curved text detection and semi- or weakly-supervised text detection.

\subsection{Curved scene text detection}
In \cite{yuliang2017detecting}, a curve text detector (CTD) integrating the recurrent transverse and longitudinal offset connection (TLOC)
is proposed to regress the relative coordinates of polygons points.
In \cite{ch2017total}, DeconvNet \cite{noh2015learning} is used to perform curved text detection.
Textsnake \cite{long2018textsnake} models scene text as a sequence of ordered, overlapping disks centered at symmetric axes,
 each of which is associated with
potential variable radius and orientations. This method performs well on arbitrary shape text but is slowed down by a complex post processing.
Mask TextSpotter \cite{lyu2018mask} directly utilizes Mask R-CNN \cite{he2017mask} to detect text and a character sensitive
segmentation branch is added as a parallel branch to original mask branch, which leads to end-to-end text spotting for arbitrary shape scene text.
FTSN \cite{dai2018fused} adapts FCIS \cite{li2017fully} for irregular text detection. All these methods are based on fully pixel-level
annotated data.

\subsection{Semi- and weakly-supervised scene text detection}
Li et al. \cite{rong2017weakly} present a weakly supervised method to generate text proposals by generating
class activation map with only image label annotations. MSER \cite{neumann2010method} is extracted from probable text regions and aggregated as groups.
Wetext \cite{tian2017wetext} propose a weakly supervised framework to train a robust character detector with a small amount of fully annotated character level
data and a large amount unannotated data or weakly annotated data with word level bounding boxes.
The word bounding boxes carry high level semantic information thus can be used to guide the semi-supervised process.
A graph-based method \cite{tian2015text} is used to group characters into words. This work performs well on horizontal text but could not be
extended to multi-oriented or arbitrary shape situation.
Wordsup \cite{hu2017wordsup} also takes character as basic elements and explores more annotations
to train a robust character detector. Through text structure analysis, this model can process several kinds of scene text like math expressions.
SEE \cite{bartz2018see} utilizes STN \cite{jaderberg2015spatial} to achieve end-to-end text spotting without the annotations of text localization.
To the best of our knowledge, semi- and weakly-supervised text detection based on segmentation is still an open problem.

\section{Proposed Methodology}
\label{sec3}
\subsection{Basic detection model}
\label{sec3.1}
Inspired by Mask R-CNN \cite{he2017mask}, we build a scene text detector that is able to localize curved text accurately.
The whole architecture is illustrated in Figure \ref{struc}.
The framework consists of five modules: feature extractor, text proposal network, region feature sampler, coarse localization network
and fine localization network. Text proposal network generates text proposals based on features extracted by the feature extractor.
Given extracted convolution features and text proposals, region features are sampled by region feature sampler and then used as inputs of
coarse localization network and fine localization network to produce outputs. During inference, the text proposals are refined by the coarse localization network first.
Given the refined boxes, the features are sampled through region feature sampler and then fed into fine localization network to generate masks.

\par\noindent\textbf{Feature Extractor} is made up of standard ResNet50 \cite{he2016deep} and FPN \cite{lin2017feature}.
FPN is top-down architecture to build multi-scale feature maps with high-level semantics.

\par\noindent\textbf{Text Proposal Network} made up of a fully convolutional network is used to generate text proposals.
We adapt the anchors of aspect ratios $\{0.2, 0.5, 1, 2, 5\}$ on five FPN stages $\{P_2, P_3, P_4, P_5, P_6\}$ to better
fit the shape and size of scene text.

\par\noindent\textbf{Region Feature Sampler} sampled region features through ROI-Align \cite{he2017mask} which have an accurate map between the
original feature maps and sampled feature maps. Region features of spatial shape of 7$\times$7 and 14$\times$14 are sampled for
coarse localization network and fine localization network respectively.

\par\noindent\textbf{Coarse Localization Network} consisting of two fully-connected layers is to perform text/non-text classification and
coarsely localize curved text with rectangles.

\par\noindent\textbf{Fine Localization Network} made up of four convolution layers is to localize curved text finely with binary mask within given rectangles.

\subsection{Optimization}
\label{optimization}
\par\noindent\textbf{Loss Function} Different from usual cascaded structure, the coarse localization network and
fine localization network is parallel in training.
The prediction target for fine localization network is the intersection between an ROI and its associated ground-truth mask.
The multi-task loss is defined as follows:
\begin{equation}
L = L_{TPN} + \lambda_{1}L_{CLN} + \lambda_{2}L_{FLN}
\end{equation}
$L_{TPN}$ and $L_{CLN}$ are the loss functions of RPN \cite{ren2015faster} and Fast R-CNN \cite{girshick2015fast}. And $L_{FLN}$ is a binary cross-entropy loss.
 $\lambda_{1}$ and $\lambda_{2}$ are balancing parameters and are both set to 1 in our experiment.

\subsection{Learning strategies}
\label{strategies}
We investigate three learning strategies including one semi-supervised strategy and two weakly-supervised strategies to deal with the
situation in practice where usually only a small amount of strong annotated data are available.
Both semi- and weakly-supervised learning strategies here assume a small amount of pixel-level annotated data.
A baseline model is first obtained with pixel-level annotated data, then are expected to be improved with a large amount of unannotated data
or weakly annotated data with horizontal rectangles.
The details of the three strategies are described in the following three subsections.

\subsubsection{Naive strategy}
\label{naive}
In semi-supervised situation, we expect to improve a scene text detector learning from unannotated data.
First, we obtain a baseline model $M$ by training with a small amount of pixel-level annotated data $D$.
Then we apply $M$ to a large amount of unannotated data $U$ and
get a candidate pseudo annotation set $C = \{(b_1, m_1, s_1), ..., (b_i, m_i, s_i), ..., (b_n, m_n, s_n)\}$,
where $(b_i, m_i, s_i)$ corresponds to bounding box, binary mask and confidence score of the i-th detection results respectively.
The final pseudo annotation set $P$ is constructed by thresholding confidence score:
\begin{equation}
P = \{(b_i, m_i)| s_i > S, (b_i, m_i, s_i) \in C\}
\end{equation}
where $S$ is the confidence threshold that is used to select positive pseudo annotations.
The pseudo annotation set $P$ together with the corresponding unannotated data are added to the fully annotated data $D$ to retrain a better model.

\subsubsection{Filter strategy}
\label{filter}
The weakly supervised learning in our scheme is to improve the performance of scene text detection model by using a large amount of
weakly annotated data $W$ with horizontal rectangles $G=\{g_1, ..., g_j, ..., g_n\}$.
The horizontal rectangles carry prior information that the areas far away from these rectangles have no text.
Similar to naive semi-supervised strategy described in last subsection, the baseline model $M$ is first applied to the weakly annotated data $W$
and a candidate pseudo annotation set $C$ is obtained.
Inspired by Wetext \cite{tian2017wetext}, we propose a strategy to filter out the false positives according to the IOU between the
detection results and the ground-truth bounding boxes. The final pseudo annotation set $P$ is constructed as follows:
\begin{equation}
\begin{aligned}
P = \{ (b_i, m_i) | &s_i > S^{'}, (b_i, m_i, s_i) \in C ,\\
&\max_{j} IOU(b_i, g_j) > T, g_j \in G\}
\end{aligned}
\end{equation}
where $S^{'}$ is the confidence score and $T$ is the IOU threshold to select positive pseudo annotations.
Also the pseudo annotation set $P$ together with the corresponding weakly annotated data are added to
the fully annotated data $D$ to retrain a better model.


\begin{figure}[t]
	\centering
	\includegraphics[width=0.9\linewidth]{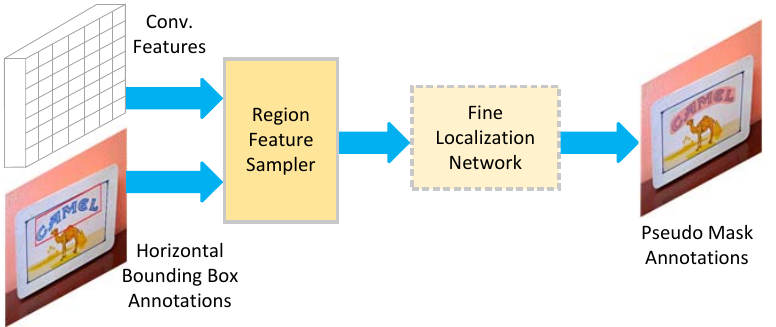}
	\caption{Pseudo annotations generation of the local strategy.}
	\label{local_flow}
\end{figure}

\subsubsection{Local strategy}
\label{local}
Different from the filter strategy in which the model is taken as a whole, the local strategy analyzes the inner structure of
the scene text detector and the relationship between a bounding box and the mask of the corresponding instance.
The fine localization network learns a map between the region feature and the instance mask.
Instead of using the proposals produced by the text proposal network, we directly take the ground-truth bounding boxes as proposals.
The region features corresponding to the proposals are sampled through the region feature sampler and then fed into
the fine localization network to generate pseudo masks.
The process of generating pseudo masks can be illustrated in Figure \ref{local_flow}.
Each rectangle and corresponding pseudo mask is taken as positive pseudo annotation:
\begin{equation}
P = \{(g_i, m_i)|g_i \in G\}
\end{equation}
Also a better model is retrained.

\subsection{Recursive Training}
\label{recursive_training}
After each training round, a better detection model can be obtained and can annotate the unannotated or weakly annotated data more precisely.
As a result, recursive training is adopted in semi- and weakly-supervised learning to improve performance.

\section{Experiments}
\label{sec4}
To evaluate the effectiveness of our proposed method, we perform experiments on two curved text datasets: CTW1500 and Total-Text.

\subsection{Datasets}
\label{sec4.1}
\par\noindent\textbf{CTW1500} is a curved text dataset which includes 10751 text annotations in 1500 images.
1000 images are selected as training set and other 500 images are taken as test set.
Each text instance is annotated in text line level with a 14 points polygon.

\par\noindent\textbf{Total-Text} contains 1555 scene images, of which 1255 training images for training and 300 images for test.
This dataset is made up of horizontal, multi-oriented and curved text. 4625 out of 9330 total text instances are curved texts.
Annotations are given in word level with polygons.

For both CTW1500 and Total-Text, we randomly select 10\% of original training images as fully annotated data and take other 90\%
as unannotated or weakly annotated data, resulting in a 100 and 900 split for CTW1500, and a 125 and 1130 split for Total-Text.

\subsection{Implementation Details}
\label{sec4.2}
We implement our work based on mmdetetion \cite{mmdetection2018}.
Similar to the original Mask R-CNN, we also fine-tune from a pre-trained ResNet50 model with initial learning rate 0.02, momentum 0.9, weight decay 0.0005.
A warmup strategy \cite{goyal2017accurate} is used in initial 500 iterations. The batch size is set to 4.
In addition, images is resized to 1333$\times$800 during both training and inference.
The score thresholds $S$ and $S^{'}$ are empirically set at 0.5 and 0.4, the IOU threshold $T$ in filter strategy is set to 0.3.

We evaluate four settings on the two datasets.
The first setting is fully supervised learning, in which the whole fully annotated training data are used to train a model.
This fully supervised setting give an upper bound,
and it could be used to verify the effectiveness of the proposed strategies.
The other three settings are naive strategy, filter strategy and local strategy which are described in section \ref{strategies}.

After training a baseline model, the four settings follow the same training settings.
At each training round, the initial baseline model is fine-tuned 60 epochs with the combination of the pixel-level annotated data
and the pseudo annotated data in the latest round.

\begin{figure}[!t]
	\centering
	\includegraphics[width=0.8\linewidth]{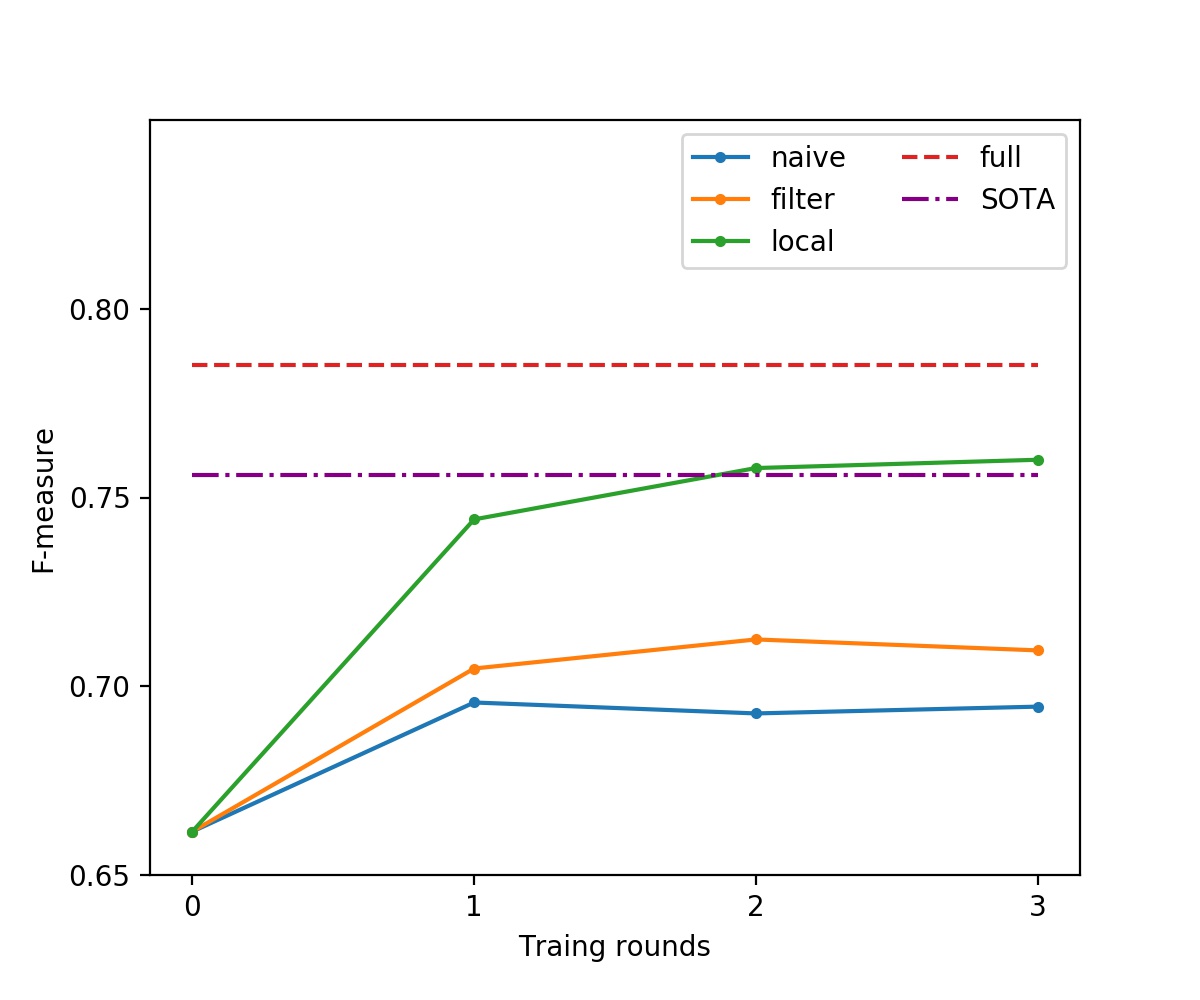}
	\caption{F-measure versus training round for different strategies on CTW1500 test set. SOTA here corresponds to \cite{long2018textsnake}}.
	\label{recu_ctw}
\end{figure}
\begin{figure}[!t]
	\centering
	\includegraphics[width=0.8\linewidth]{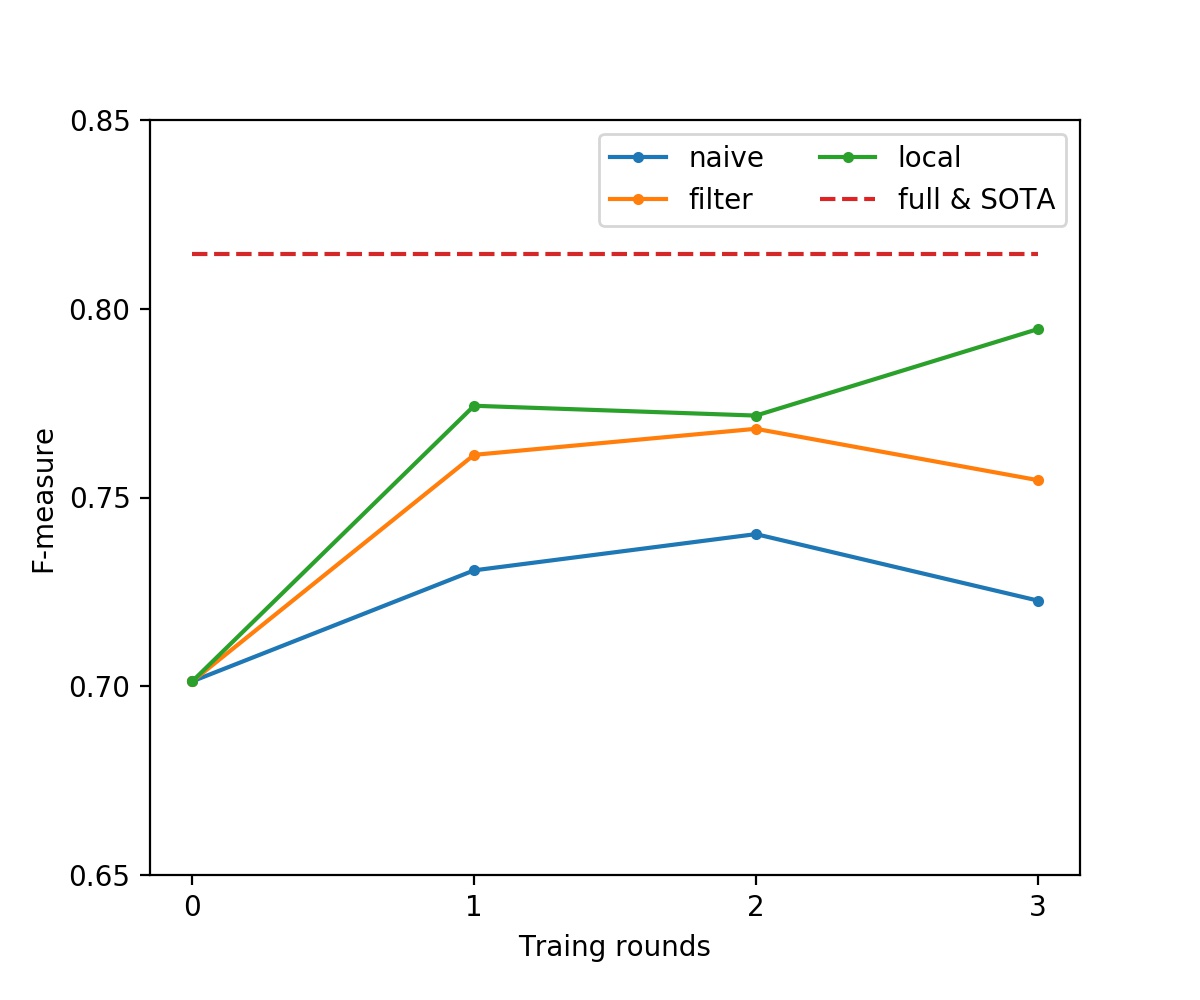}
	\caption{F-measure versus training round for different strategies on Total-Text test set. SOTA here corresponds to \cite{dai2018fused}}.
	\label{recu_total}
\end{figure}





\subsection{Experimental Results}
\label{sec4.3}

\begin{table}[b]
\small
\centering
\caption{Detection results on CTW1500 test set(\%), \protect\\ * indicates result from \cite{yuliang2017detecting}}
\label{tab1}
\begin{tabular}{|c|c|c|c|}
		\hline
		{\bf Method}&{\bf Precision}&{\bf Recall}&{\bf F-measure}\\
		\hline
		SegLink * \cite{shi2017detecting}&42.3&40.0&40.8\\
		\hline
		CTPN * \cite{tian2016detecting}&60.4&53.8&56.9\\
		\hline
		EAST * \cite{zhou2017east}&{\bf 78.7}&49.1&60.4\\
		\hline
		DMPNet * \cite{liu2017deep}&69.9&56.0&62.2\\
		\hline
		CTD \cite{yuliang2017detecting}&74.3&65.2&69.5\\
		\hline
		CTD+TLOC \cite{yuliang2017detecting}&77.4&69.8&73.4\\
		\hline
		TextSnake \cite{long2018textsnake}&67.9&{\bf 85.3}&75.6\\
		\hline
		{\bf Baseline}&61.8&71.0&66.1\\
		\hline
		{\bf Naive}&66.9&72.3&69.5\\
		\hline
		{\bf Filter}&74.7&68.0&71.2\\
		\hline
		{\bf Local}&73.8&78.2&{\bf 76.0}\\
		\hline
		{\bf Fully}&77.0&79.9&{\bf 78.5}\\
		\hline
\end{tabular}
\end{table}

\begin{table}[!htbp]
\small
\centering
\caption{Detection results on Total-Text test set(\%), \protect\\ * indicates result from \cite{long2018textsnake}}
\label{tab2}
\begin{tabular}{|c|c|c|c|}
		\hline
		{\bf Method}&{\bf Precision}&{\bf Recall}&{\bf F-measure}\\
		\hline
		SegLink * \cite{shi2017detecting}&30.3&23.8&26.7\\
		\hline
		EAST * \cite{zhou2017east}&50.0&36.2&42.0\\
		\hline
		DeconvNet \cite{ch2017total}&33.0&40.0&36.0\\
		\hline
        Mask TextSpotter \cite{lyu2018mask}&69.0&55.0&61.3\\
        \hline
		TextSnake \cite{long2018textsnake}&82.7&74.5&78.4\\
		\hline
		FTSN \cite{dai2018fused}&{\bf 84.7}&78.0&81.3\\
		\hline
		{\bf Baseline}&65.2&75.8&70.1\\
		\hline
		{\bf Naive}&71.0&77.3&74.0\\
		\hline
		{\bf Filter}&79.8&73.9&76.8\\
		\hline
		{\bf Local}&77.5&{\bf 81.5}&79.4\\
		\hline
		{\bf Fully}&80.6&{\bf 82.3}&{\bf 81.4}\\
		\hline
\end{tabular}
\end{table}


Performance in all three strategies is found to be improved in recursive training.
Figure \ref{recu_ctw} and Figure \ref{recu_total} present the results of recursive training of three strategies within three training rounds.
The result of the 0 training round is the performance of the baseline model.
The state-of-the-art method is also presented for reference.
In Figure \ref{recu_ctw}, the fully supervised model outperform the state-of-the-art method \cite{long2018textsnake} in that
the former method's backbone is ResNet50 whereas the latter method's backbone is VGG16.
We can observe the improvement of performance within several training rounds.
Naive strategy and filter strategy achieve top performance in first two training rounds.
However, local strategy achieves top performance in the third training round. The fast saturation for the naive and filter strategies maybe
originate from the large noise rates of the pseudo annotated data. The slow saturation also demonstrates the superiority of
the proposed local strategy.

\begin{figure*}[t]


\begin{minipage}[b]{.19\linewidth}
  \centering
  \centerline{\epsfig{figure=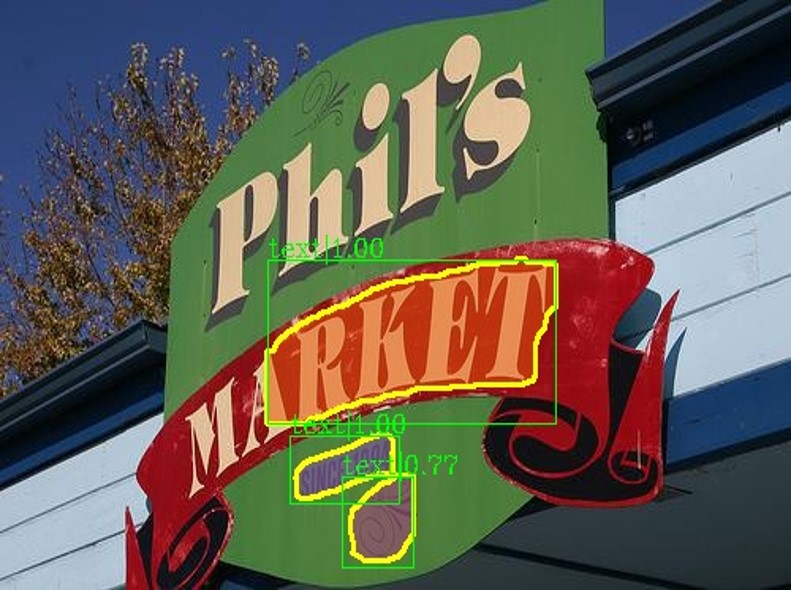,width=3.3cm}}
  \vspace{0.1cm}
\end{minipage}
\begin{minipage}[b]{.19\linewidth}
  \centering
  \centerline{\epsfig{figure=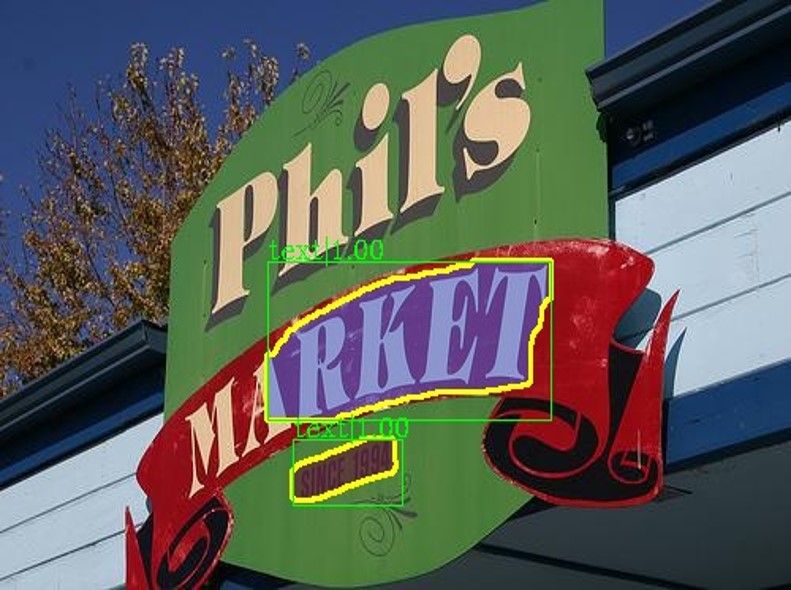,width=3.3cm}}
  \vspace{0.1cm}
\end{minipage}
\begin{minipage}[b]{.19\linewidth}
  \centering
  \centerline{\epsfig{figure=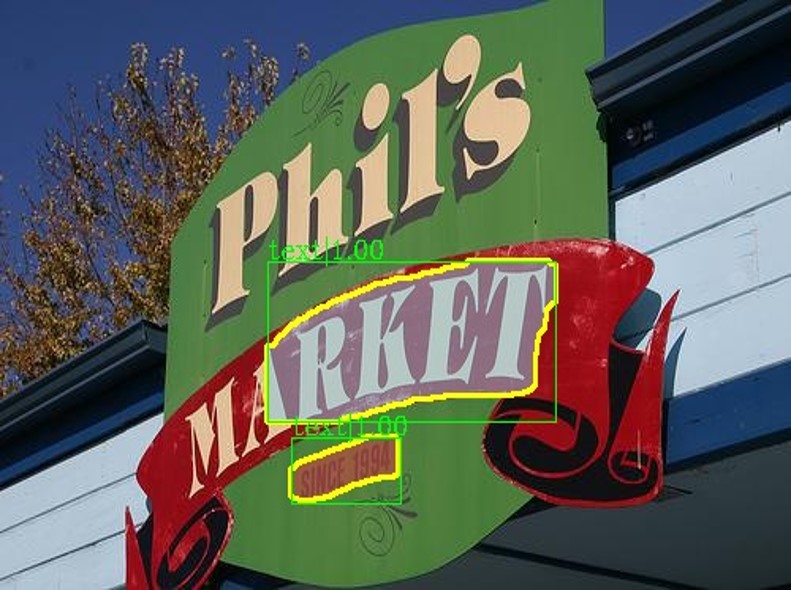,width=3.3cm}}
  \vspace{0.1cm}
\end{minipage}
\begin{minipage}[b]{.19\linewidth}
  \centering
  \centerline{\epsfig{figure=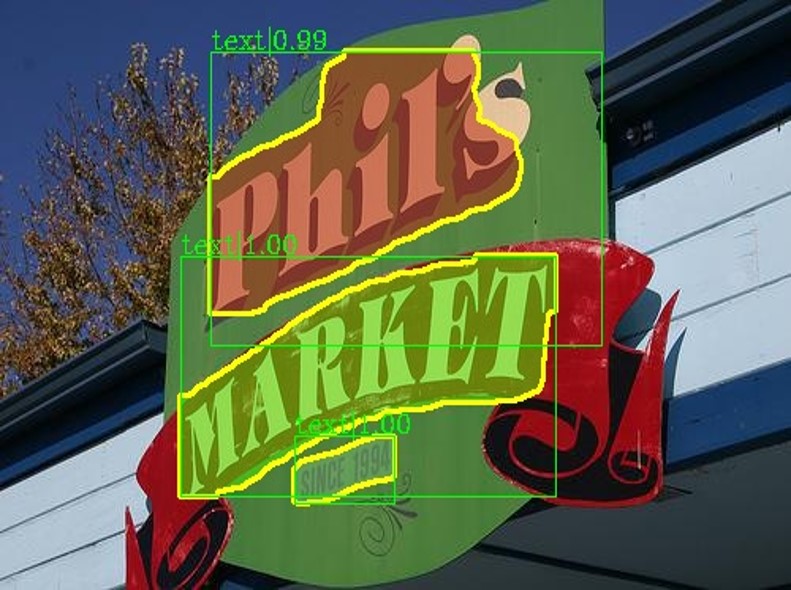,width=3.3cm}}
  \vspace{0.1cm}
\end{minipage}
\begin{minipage}[b]{.19\linewidth}
  \centering
  \centerline{\epsfig{figure=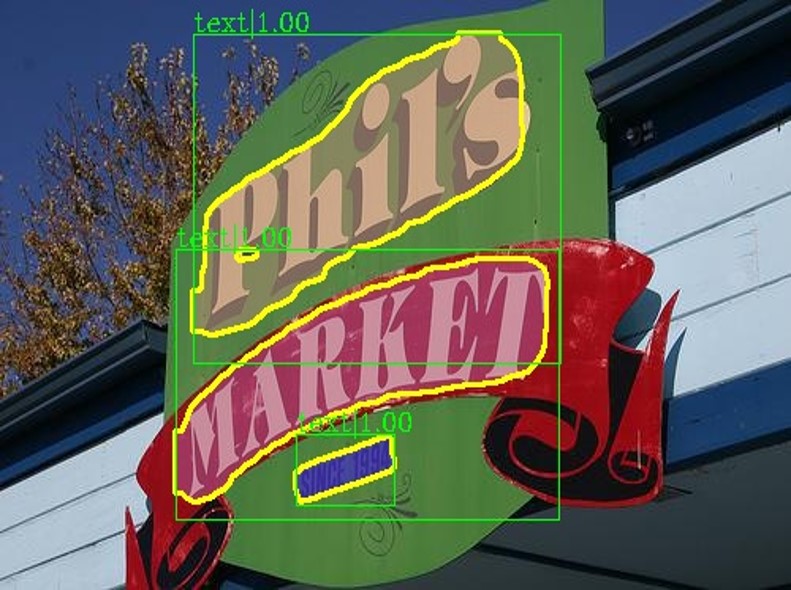,width=3.3cm}}
  \vspace{0.1cm}
\end{minipage}


\begin{minipage}[b]{.19\linewidth}
  \centering
  \centerline{\epsfig{figure=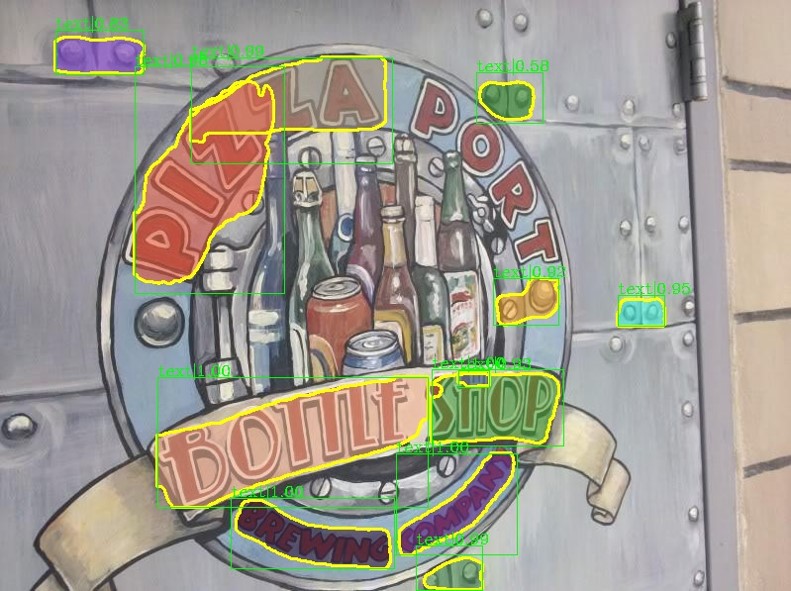,width=3.3cm}}
  \vspace{0.1cm}
\end{minipage}
\begin{minipage}[b]{.19\linewidth}
  \centering
  \centerline{\epsfig{figure=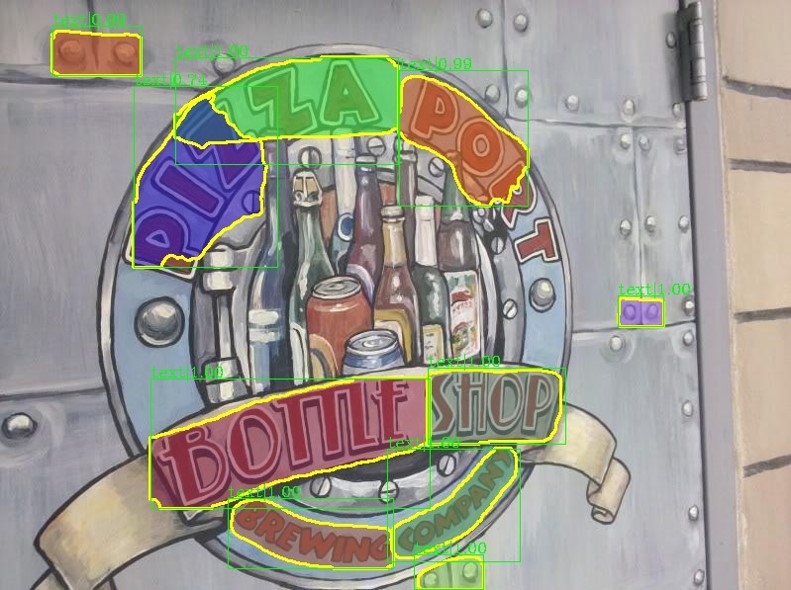,width=3.3cm}}
  \vspace{0.1cm}
\end{minipage}
\begin{minipage}[b]{.19\linewidth}
  \centering
  \centerline{\epsfig{figure=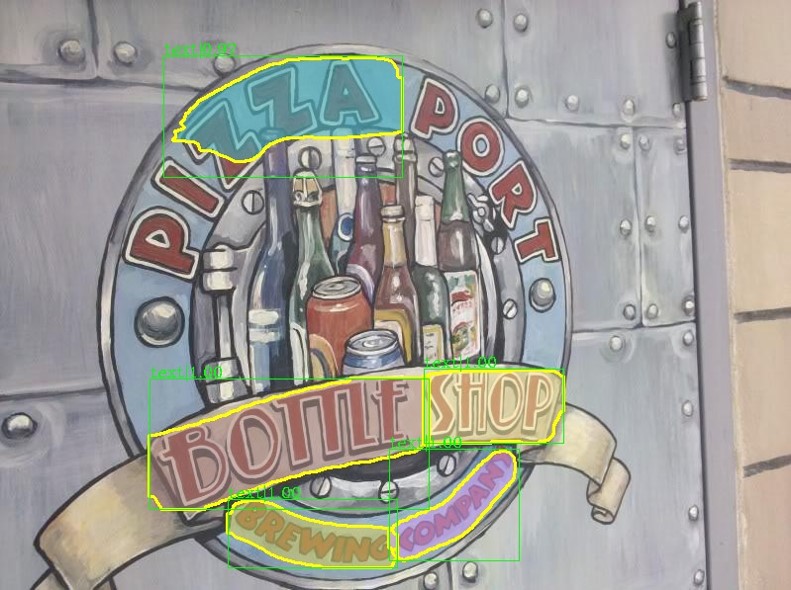,width=3.3cm}}
  \vspace{0.1cm}
\end{minipage}
\begin{minipage}[b]{.19\linewidth}
  \centering
  \centerline{\epsfig{figure=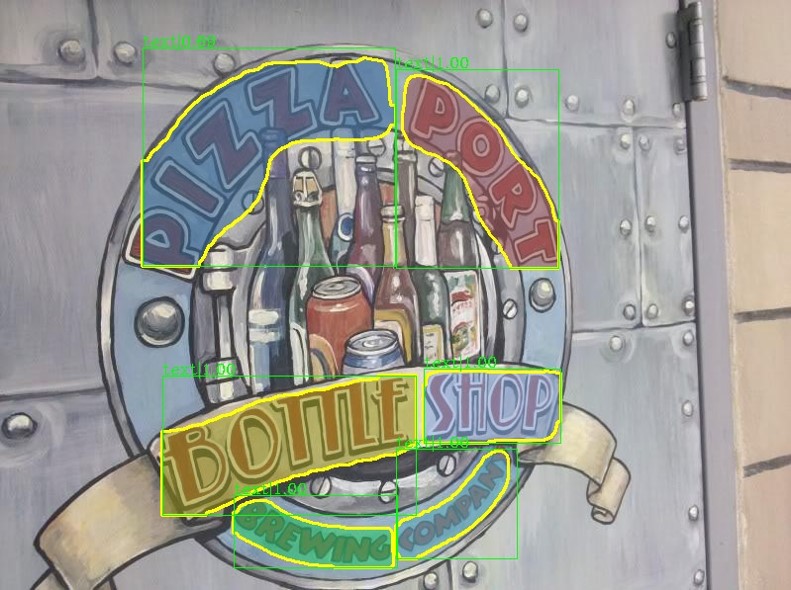,width=3.3cm}}
  \vspace{0.1cm}
\end{minipage}
\begin{minipage}[b]{.19\linewidth}
  \centering
  \centerline{\epsfig{figure=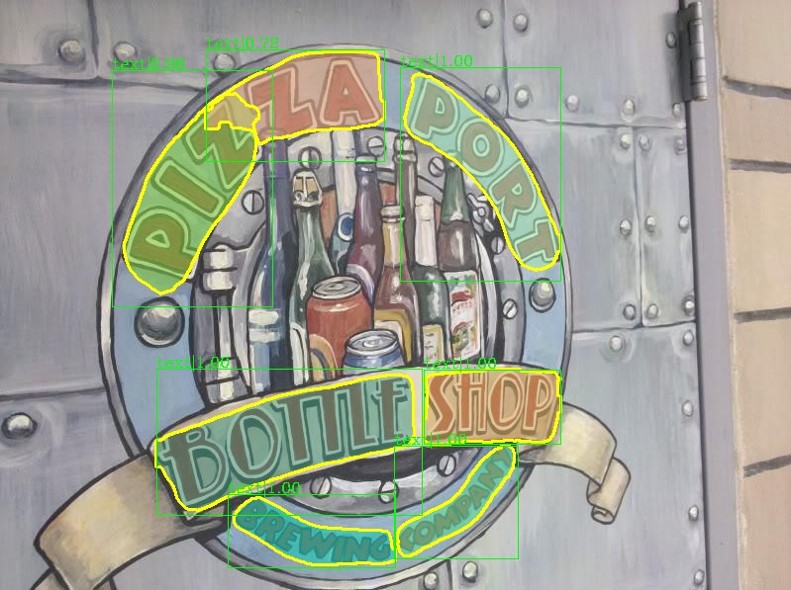,width=3.3cm}}
  \vspace{0.1cm}
\end{minipage}

\caption{Comparison of detection results on CTW1500 in the first row and Total-Text in the last row.
Five columns are results of baseline, naive strategy, filter strategy, local strategy and fully supervised situation respectively.}
\label{Comparison}
\end{figure*}

The results on CTW1500 and Total-Text are listed in Table \ref{tab1} and Table \ref{tab2} respectively.
The model with the top performance across all training rounds is picked for evaluation.
On CTW1500, the proposed three strategies outperform the baseline model by 3.4\%, 5.1\% and 9.9\% respectively.
In naive strategy, the precision is higher than the baseline while the recall rate is maintained.
The filter strategy outperforms the naive strategy by 1.7\%, and the higher precision rate means it can remove false positives.
The local strategy outperforms the filter strategy by 4.8\% F-score.
The high recall rate benefits from the fully supervision of the coarse localization network.
The local strategy is comparable with or even superior to the state-of-the-art method with only 10\% pixel-level annotated data and 90\%
weakly annotated data.On Total-Text, the three strategies outperform the baseline model by 3.9\%, 6.7\%, 9.3\% F-measure.
The local strategy also perform best in these three strategies.
These results are consistent with the results on CTW1500. Specially, the local strategy achieves state of the art recall rate.
Comparison results are showed in Figure \ref{Comparison}. The local strategy outperforms the other two strategies and is close to
fully supervised detection results.

\begin{figure}[!t]
\begin{minipage}[b]{0.48\linewidth}
	\centering
	\centerline{\epsfig{figure=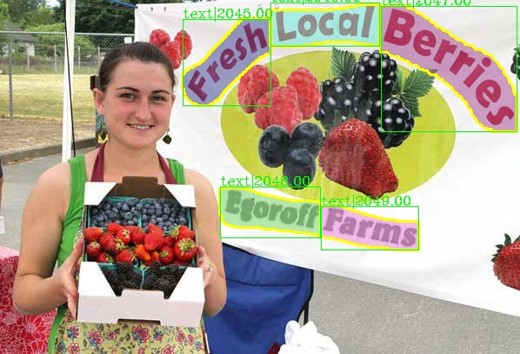,width=4.0cm}}
\end{minipage}
\begin{minipage}[b]{0.48\linewidth}
	\centering
	\centerline{\epsfig{figure=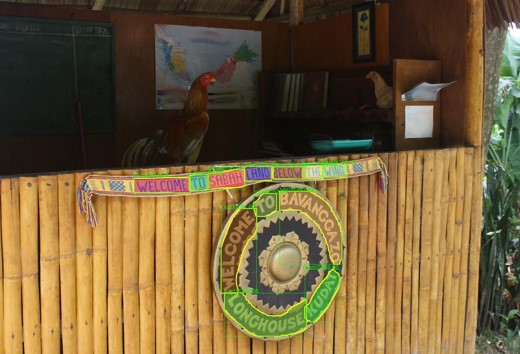,width=4.0cm}}
\end{minipage}

\caption{Annotation results on Total-Text with local strategy using the model trained on CTW1500.}
\label{generation}
\end{figure}

Except the experiments on CTW1500 and Total-Text, we experiment on a more challenging task across different datasets.
We assume a scenario that there are a small amount of fully annotated source data and a large amount of weakly annotated source data with
bounding box annotations, and there are a large amount of target data which is weakly annotated with horizontal bounding box.
The target is to learn a robust detector that can work on the target domain.
This task is very challenging because different datasets usually have different distributions.
In our experiment, CTW1500 is used as the source data and the split of 100 and 900 is taken as pixel-level annotated data
and weak annotated data with rectangles.
Total-Text is chosen as the target data and the whole training set including 1255 images are used as weakly annotated target data.
We use this setting because that CTW1500 is annotated in text line level and Total-Text is annotated in word level.
Our local strategy is suitable for this task. We use the best model of the local strategy on CTW1500 that achieve 76.0\% on CTW1500 test set.
We use the model to annotate Total-Text data in the same manner of local strategy.
The model is fine-tuned with pseudo annotated data of Total-Text. Recursive training is also used.
78.1\% F-measure is achieved in such a scenario and is comparable to TextSnake which achieves 78.4\% F-measure.
Figure \ref{generation} shows the annotation results on Total-Text with local strategy using the model trained on CTW1500.
These demonstrate good generalization of the proposed method.



\subsection{Discussion}
\label{sec4.4}
WeText may be the most relative work to our method.
Both works focus on how to use a small fully annotated data to train a robust and accurate model utilizing semi-supervised
and weakly supervised learning. WeText is designed to detect character, and exploits the relationship between character bounding box and
word bounding box.
However, our method is directly designed for arbitrary word/text line detection and
exploits relationship between bounding box and polygon/pixel-level annotation.

\section{Conclusion}
In this work, we propose a novel semi- and weakly-supervised framework for curved text detection.
The framework substantially reduces requirement of pixel-level annotations.
Based on a baseline model that is trained on a small amount of pixel-level annotated data,
three strategies are adopted to annotate the large amount of weakly annotated or unannotated data.
Experiments show that the local strategy outperforms other strategies.
The proposed method is comparable with state-of-the-art methods, and also generalize well across different datasets.
In the future, we will explore the relative work on weakly supervised object detection \cite{8640243} and expect these works
will promote our weakly supervised text detection work.

\section*{Acknowledgment}
This work is supported by the National Key R\&D Program of China (2017YFB1002400)
and the Strategic Priority Research Program of Chinese Academy of Sciences (XDC02000000).

\scriptsize
\bibliographystyle{IEEEtran}
\bibliography{references}

\begin{thebibliography}{10}
\providecommand{\url}[1]{#1}
\csname url@samestyle\endcsname
\providecommand{\newblock}{\relax}
\providecommand{\bibinfo}[2]{#2}
\providecommand{\BIBentrySTDinterwordspacing}{\spaceskip=0pt\relax}
\providecommand{\BIBentryALTinterwordstretchfactor}{4}
\providecommand{\BIBentryALTinterwordspacing}{\spaceskip=\fontdimen2\font plus
\BIBentryALTinterwordstretchfactor\fontdimen3\font minus
  \fontdimen4\font\relax}
\providecommand{\BIBforeignlanguage}[2]{{%
\expandafter\ifx\csname l@#1\endcsname\relax
\typeout{** WARNING: IEEEtran.bst: No hyphenation pattern has been}%
\typeout{** loaded for the language `#1'. Using the pattern for}%
\typeout{** the default language instead.}%
\else
\language=\csname l@#1\endcsname
\fi
#2}}
\providecommand{\BIBdecl}{\relax}
\BIBdecl

\bibitem{ye2015text}
Q.~Ye and D.~Doermann, ``Text detection and recognition in imagery: A survey,''
  \emph{TPAMI}, vol.~37, no.~7, pp. 1480--1500, 2015.

\bibitem{zhu2016scene}
Y.~Zhu, C.~Yao, and X.~Bai, ``Scene text detection and recognition: Recent
  advances and future trends,'' \emph{Frontiers of Computer Science}, vol.~10,
  no.~1, pp. 19--36, 2016.

\bibitem{long2018scene}
S.~Long, X.~He, and C.~Ya, ``Scene text detection and recognition: The deep
  learning era,'' \emph{arXiv preprint arXiv:1811.04256}, 2018.

\bibitem{karatzas2013icdar}
D.~Karatzas, F.~Shafait, S.~Uchida, M.~Iwamura, L.~G. i~Bigorda, S.~R. Mestre,
  J.~Mas, D.~F. Mota, J.~A. Almazan, and L.~P. De~Las~Heras, ``{ICDAR} 2013
  robust reading competition,'' in \emph{ICDAR}, 2013, pp. 1484--1493.

\bibitem{karatzas2015icdar}
D.~Karatzas, L.~Gomez-Bigorda, A.~Nicolaou, S.~Ghosh, A.~Bagdanov, M.~Iwamura,
  J.~Matas, L.~Neumann, V.~R. Chandrasekhar, S.~Lu \emph{et~al.}, ``{ICDAR}
  2015 competition on robust reading,'' in \emph{ICDAR}, 2015, pp. 1156--1160.

\bibitem{ch2017total}
C.~K. Ch'ng and C.~S. Chan, ``Total-text: A comprehensive dataset for scene
  text detection and recognition,'' in \emph{ICDAR}, vol.~1, 2017, pp.
  935--942.

\bibitem{yuliang2017detecting}
L.~Yuliang, J.~Lianwen, Z.~Shuaitao, and Z.~Sheng, ``Detecting curve text in
  the wild: New dataset and new solution,'' \emph{arXiv preprint
  arXiv:1712.02170}, 2017.

\bibitem{long2018textsnake}
S.~Long, J.~Ruan, W.~Zhang, X.~He, W.~Wu, and C.~Yao, ``Textsnake: A flexible
  representation for detecting text of arbitrary shapes,'' in \emph{ECCV},
  2018, pp. 19--35.

\bibitem{dai2018fused}
Y.~Dai, Z.~Huang, Y.~Gao, Y.~Xu, K.~Chen, J.~Guo, and W.~Qiu, ``Fused text
  segmentation networks for multi-oriented scene text detection,'' in
  \emph{ICPR}, 2018, pp. 3604--3609.

\bibitem{lyu2018mask}
P.~Lyu, M.~Liao, C.~Yao, W.~Wu, and X.~Bai, ``Mask textspotter: An end-to-end
  trainable neural network for spotting text with arbitrary shapes,'' in
  \emph{ECCV}, 2018, pp. 67--83.

\bibitem{long2015fully}
J.~Long, E.~Shelhamer, and T.~Darrell, ``Fully convolutional networks for
  semantic segmentation,'' in \emph{CVPR}, 2015, pp. 3431--3440.

\bibitem{li2017fully}
Y.~Li, H.~Qi, J.~Dai, X.~Ji, and Y.~Wei, ``Fully convolutional instance-aware
  semantic segmentation,'' in \emph{CVPR}, 2017, pp. 4438--4446.

\bibitem{tian2017wetext}
S.~Tian, S.~Lu, and C.~Li, ``Wetext: Scene text detection under weak
  supervision,'' in \emph{ICCV}, 2017, pp. 1492--1500.

\bibitem{li2017towards}
H.~Li, P.~Wang, and C.~Shen, ``Towards end-to-end text spotting with
  convolutional recurrent neural networks,'' in \emph{ICCV}, 2017, pp.
  5238--5246.

\bibitem{noh2015learning}
H.~Noh, S.~Hong, and B.~Han, ``Learning deconvolution network for semantic
  segmentation,'' in \emph{ICCV}, 2015, pp. 1520--1528.

\bibitem{he2017mask}
K.~He, G.~Gkioxari, P.~Doll{\'a}r, and R.~Girshick, ``Mask r-cnn,'' in
  \emph{ICCV}, 2017, pp. 2980--2988.

\bibitem{rong2017weakly}
R.~Li, M.~En, J.~Li, and H.~Zhang, ``Weakly supervised text attention network
  for generating text proposals in scene images,'' in \emph{ICDAR}, vol.~1,
  2017, pp. 324--330.

\bibitem{neumann2010method}
L.~Neumann and J.~Matas, ``A method for text localization and recognition in
  real-world images,'' in \emph{ACCV}, 2010, pp. 770--783.

\bibitem{tian2015text}
S.~Tian, Y.~Pan, C.~Huang, S.~Lu, K.~Yu, and C.~Lim~Tan, ``Text flow: A unified
  text detection system in natural scene images,'' in \emph{ICCV}, 2015, pp.
  4651--4659.

\bibitem{hu2017wordsup}
H.~Hu, C.~Zhang, Y.~Luo, Y.~Wang, J.~Han, and E.~Ding, ``Wordsup: Exploiting
  word annotations for character based text detection,'' in \emph{ICCV}, 2017,
  pp. 4940--4949.

\bibitem{bartz2018see}
C.~Bartz, H.~Yang, and C.~Meinel, ``{SEE}: towards semi-supervised end-to-end
  scene text recognition,'' in \emph{AAAI}, 2018.

\bibitem{jaderberg2015spatial}
M.~Jaderberg, K.~Simonyan, A.~Zisserman \emph{et~al.}, ``Spatial transformer
  networks,'' in \emph{NeurIPS}, 2015, pp. 2017--2025.

\bibitem{he2016deep}
K.~He, X.~Zhang, S.~Ren, and J.~Sun, ``Deep residual learning for image
  recognition,'' in \emph{CVPR}, 2016, pp. 770--778.

\bibitem{lin2017feature}
T.-Y. Lin, P.~Doll{\'a}r, R.~Girshick, K.~He, B.~Hariharan, and S.~Belongie,
  ``Feature pyramid networks for object detection,'' in \emph{CVPR}, 2017, pp.
  2117--2125.

\bibitem{ren2015faster}
S.~Ren, K.~He, R.~Girshick, and J.~Sun, ``Faster r-cnn: Towards real-time
  object detection with region proposal networks,'' in \emph{NeurIPS}, 2015,
  pp. 91--99.

\bibitem{girshick2015fast}
R.~Girshick, ``Fast r-cnn,'' in \emph{ICCV}, 2015, pp. 1440--1448.

\bibitem{mmdetection2018}
K.~Chen, J.~Pang, J.~Wang, Y.~Xiong, X.~Li, S.~Sun, W.~Feng, Z.~Liu, J.~Shi,
  W.~Ouyang, C.~C. Loy, and D.~Lin, ``mmdetection,''
  \url{https://github.com/open-mmlab/mmdetection}, 2018.

\bibitem{goyal2017accurate}
P.~Goyal, P.~Doll{\'a}r, R.~Girshick, P.~Noordhuis, L.~Wesolowski, A.~Kyrola,
  A.~Tulloch, Y.~Jia, and K.~He, ``Accurate, large minibatch sgd: training
  imagenet in 1 hour,'' \emph{arXiv preprint arXiv:1706.02677}, 2017.

\bibitem{shi2017detecting}
B.~Shi, X.~Bai, and S.~Belongie, ``Detecting oriented text in natural images by
  linking segments,'' \emph{arXiv preprint arXiv:1703.06520}, 2017.

\bibitem{tian2016detecting}
Z.~Tian, W.~Huang, T.~He, P.~He, and Y.~Qiao, ``Detecting text in natural image
  with connectionist text proposal network,'' in \emph{ECCV}, 2016, pp. 56--72.

\bibitem{zhou2017east}
X.~Zhou, C.~Yao, H.~Wen, Y.~Wang, S.~Zhou, W.~He, and J.~Liang, ``{EAST}: an
  efficient and accurate scene text detector,'' in \emph{CVPR}, 2017, pp.
  2642--2651.

\bibitem{liu2017deep}
Y.~Liu and L.~Jin, ``Deep matching prior network: Toward tighter multi-oriented
  text detection,'' in \emph{CVPR}, 2017, pp. 3454--3461.

\bibitem{8640243}
F.~{Wan}, P.~{Wei}, Z.~{Han}, J.~{Jiao}, and Q.~{Ye}, ``Min-entropy latent
  model for weakly supervised object detection,'' \emph{TPAMI}, pp. 1--1, 2019.

\end{thebibliography}

\end{document}